\documentclass[10pt,twocolumn,letterpaper]{article}

\usepackage[pagenumbers]{cvpr} 

\usepackage[accsupp]{axessibility}  

\usepackage{graphicx}
\usepackage{amsmath}
\usepackage{amssymb}
\usepackage{booktabs}

\usepackage[pagebackref,breaklinks,colorlinks]{hyperref}

\usepackage{threeparttable}
\usepackage{multirow}

\usepackage{rotating}
\usepackage{adjustbox}
\usepackage{soul}
\usepackage{makecell}
\usepackage{booktabs}
\usepackage{amsmath}
\usepackage{amssymb}
\usepackage{bbding}
\usepackage{pifont}

\usepackage[capitalize]{cleveref}
\crefname{section}{Sec.}{Secs.}
\Crefname{section}{Section}{Sections}
\Crefname{table}{Table}{Tables}
\crefname{table}{Tab.}{Tabs.}

\def\cvprPaperID{6308} 
\def\confName{CVPR}
\def\confYear{2023}

\begin{document}

\title{Implicit Identity Leakage: \\ The Stumbling Block to Improving Deepfake Detection Generalization}

\author{Shichao Dong$^{1,*}$, Jin Wang$^{1,*}$, Renhe Ji$^{1,\dag}$, Jiajun Liang$^{1}$, Haoqiang Fan$^{1}$, Zheng Ge$^{1}$ \\
$^1$MEGVII Technology\\
{\tt\small \{dongshichao,wangjin,jirenhe,liangjiajun,fhq,gezheng\}@megvii.com}
}
\maketitle
{
	\renewcommand{\thefootnote}{\fnsymbol{footnote}}
    \footnotetext[1]{Equal contribution}
	\footnotetext[2]{Corresponding author}
}

\begin{abstract}
In this paper, we analyse the generalization ability of binary classifiers for the task of deepfake detection.
We find that the stumbling block to their generalization is caused by the unexpected learned identity representation on images.
Termed as the Implicit Identity Leakage, this phenomenon has been qualitatively and quantitatively verified among various DNNs.
Furthermore, based on such understanding, we propose a simple yet effective method named the ID-unaware Deepfake Detection Model to reduce the influence of this phenomenon.
Extensive experimental results demonstrate that our method outperforms the state-of-the-art in both in-dataset and cross-dataset evaluation. 
The code is available at \href{https://github.com/megvii-research/CADDM}{https://github.com/megvii-research/CADDM}.
\end{abstract}

\section{Introduction}
Recently, face-swap abusers use different face manipulation methods \cite{faceswap,deepfakes,faceshifter,face2face, faceswap} to generate fake images/videos. 
Those images/videos are then used to spread fake news, make malicious hoaxes, and forge judicial evidence, which have caused severe consequences.
In order to alleviate such situations, an increasing number of deepfake detection methods \cite{ding2020swapped,tariq2018detecting,marra2018detection,song2022face,dong2022protecting,song2022adaptive} have been proposed to filter out manipulated images/videos from massive online media resources, ensuring the filtered images/videos are genuine and reliable.


\begin{figure}[htp]
\centering
\includegraphics [width=0.45\textwidth]{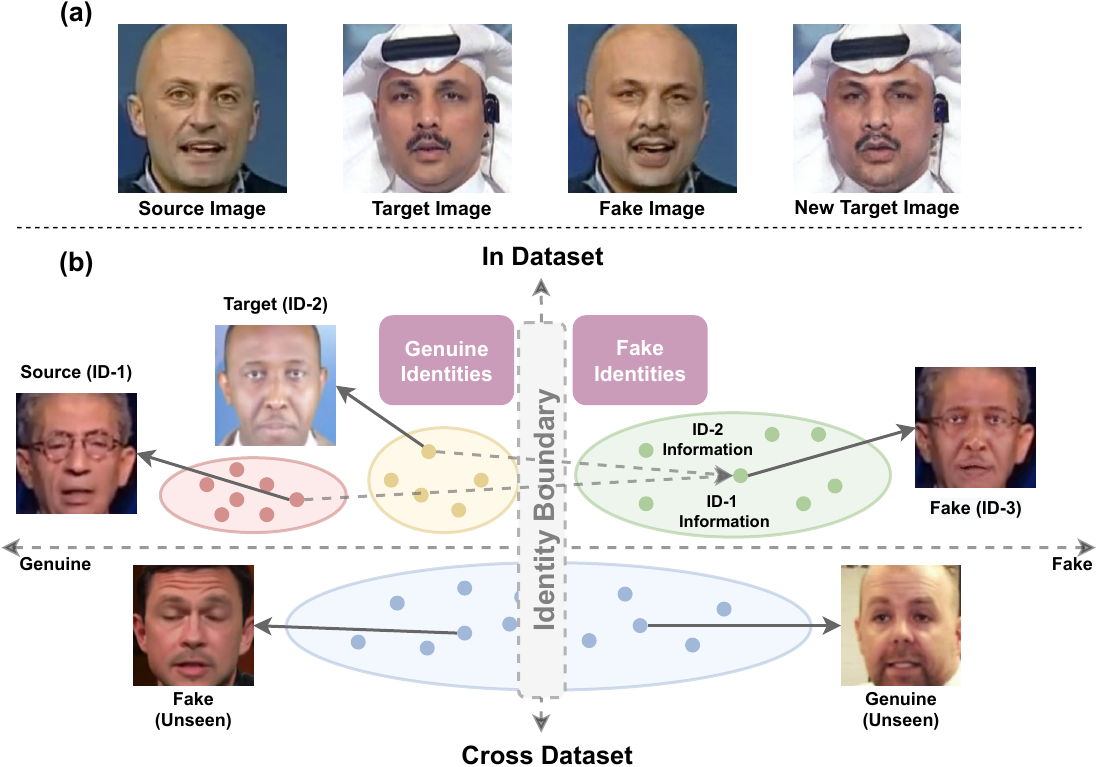}
\caption{
\textbf{The Implicit Identity Leakage phenomenon.}
Since the fake image retains some features of its source image, its identity should not be completely regarded as its target image. 
As a consequence, there exists an implicit gap between genuine identities and fake identities in the training set, which is unintentionally captured by binary classifiers. 
When confronted with images manipulated by unseen face-swap methods, the classifier tends to misuse identity information and make false predictions.
}
\label{fig:related}
\end{figure}

Previous methods usually dealt with the task of deepfake detection with binary classifiers \cite{rahmouni2017distinguishing,cozzolino2017recasting,bayar2016deep,mesonet,nguyen2019use}.
These methods have achieved great detection accuracy in detecting the seen attacks learned in the training datasets (\emph{i.e.} the in-dataset evaluations).
However, when confronted with media generated from newly-proposed deepfake methods (\emph{i.e.} the cross-dataset evaluations), these methods often suffered from significant performance drops.
Though plenty of researchers have designed effective methods \cite{face-x-ray,amazon,multi} to improve the generalization of deepfake detection models, it still lacks a thorough analysis of why binary classifiers fail to perform well on the cross-dataset evaluation.

In this paper, given well-trained binary classifiers of deepfake detection, we find that the stumbling block for their generalization ability is caused by the mistakenly learned identity representation on images. 
As shown in Fig. \ref{fig:related} (a), a deepfake image is usually generated by replacing the face of the source image with the face of the target image.
However, we notice that the procedure of synthesizing the fake image \cite{deepfakes,faceswap,simswap} may cause the information loss of ID representations. 
The identity of the fake image can not be considered as the same as either its target image or its source image.
In particular, when the face of the target image is swapped back with the face of the fake image, it is noticeable that the identity of the target image is altered.

In this way, as shown in Fig. \ref{fig:related} (b), when learning a deepfake detection model, there exists an implicit decision boundary between fake images and genuine images based on identities. 
During the training phase, binary classifiers may accidentally consider certain groups of identities as genuine identities and other groups of identities as fake identities.
When tested on the cross-dataset evaluation, such biased representations may be mistakenly used by binary classifiers, causing false judgments based on the facial appearance of images.
In this paper, we have qualitatively and quantitatively verified this phenomenon (termed as the Implicit Identity Leakage) in binary classifiers of various backbones. 
Please see Sec. \ref{sec:IIL} and Sec. \ref{sec:IIL2} for analyses.


Furthermore, based on such understanding, we propose a simple yet effective method named the ID-unaware Deepfake Detection Model to reduce the influence of Implicit Identity Leakage.
Intuitively, by forcing models to only focus on local areas of images, less attention will be paid to the global identity information.
Therefore, we design an anchor-based detector module termed as the Artifact Detection Module to guide our model to focus on the local artifact areas. 
Such a module is expected to detect artifact areas on images with multi-scale anchors, each of which is assigned a binary label to indicate whether the artifact exists.
By localizing artifact areas and classifying multi-scale anchors, our model learns to distinguish the differences between local artifact areas and local genuine areas at a finer level, thus reducing the misusage of the global identity information.

Extensive experimental results show that our model accurately predicted the position of artifact areas and learned generalized artifact features in face manipulation algorithms, successfully outperforming the state-of-the-art.
Contributions of the paper are summarized as follows: 
\begin{itemize}
\item We discover that deepfake detection models supervised only by binary labels are very sensitive to the identity information of the images, which is termed as the Implicit Identity Leakage in this paper.
\item We propose a simple yet effective method termed as the ID-unaware Deepfake Detection Model to reduce the influence of the ID representation, successfully outperforming other state-of-the-art methods.
\item We conduct extensive experiments to verify the Implicit Identity Leakage phenomenon and demonstrate the effectiveness of our method.
\end{itemize}


\section{Related Work}

With the development of Generative Adversarial Network (GAN) \cite{gan,stgan,stylegan,analystylegan,stargan} techniques, forgery images/videos have become more realistic and indistinguishable. 
To deal with attacks based on different face manipulation algorithms, researchers tried to improve their deepfake detectors \cite{quan2018distinguishing,mo2018fake,hsu2018learning} from different perspectives, such as designing different loss functions \cite{bondi2020training}, extracting richer features \cite{xuan2019generalization,du2019towards}, and analyzing the continuity between consecutive frames \cite{nirkin2021deepfake,hernandez2020deepfakeson}.
Most of these deepfake detection methods can be roughly summarized into two categories.

\subsection{Binary Classifiers}
Many researchers \cite{rahmouni2017distinguishing,cozzolino2017recasting,bayar2016deep,mesonet,nguyen2019use} treated the deepfake detection task as a binary classification problem.
They used a backbone encoder to extract high-level features and a classifier to detect whether the input image has been manipulated. 
Durall \emph{et al.} \cite{durall2019unmasking} first proposed a model analyzing the frequency domain for face forgery detection.
Masi \emph{et al.} \cite{masi2020two} used a two-branch recurrent network to extract high-level semantic information in original RGB images and their frequency domains at the same time, by which the model achieved good performance on multiple public datasets. 
Li \emph{et al.} \cite{single-center-loss} designed a single-center loss to compress the real sample classification space to further improve the detection rate of forged samples.
Binary classifiers achieved high detection accuracy on in-dataset evaluation, but they could not maintain good performance when facing unseen forged images.

\subsection{Hand-crafted Deepfake Detectors}
Many works attempted to improve the generalization capability of deepfake detectors by modeling specific hand-crafted artifacts among different face manipulation methods. 
Li \emph{et al.} \cite{li2018ictu} believed that some physical characteristics of a real person cannot be manipulated in fake videos.
They designed an eye blinking detector to identify the authenticity of the video through the frequency of eye blinking.  
Since 3D data can not be reversely generated from the fake image, Yang \emph{et al.} \cite{yang2019exposing} did the face forgery detection task from the perspective of non-3D projection generation samples.
Sun \emph{et al.} \cite{improving} and Li \emph{et al.} \cite{face-x-ray} focused on precise geometric features (face landmark) and blending artifacts respectively when detecting forged images.
Liu \emph{et al.} \cite{spatial} equipped the model with frequency domain information since the frequency domain is very sensitive to upsampling operations (which are often used in deepfake detection models), and used a shallow network to extract rich local texture information, enhancing the model’s generalization and robustness.

In summary, hand-crafted deepfake detectors guided the model to capture specific artifact features and indicated manipulated images/videos by responding to these features.
However, these methods have a common limitation: when forgeries do not contain specific artifacts that are introduced in the training phase, they often fail to work well.

\section{Implicit Identity Leakage}
\label{sec:IIL}
The Implicit Identity Leakage denotes that the ID representation in the deepfake dataset is captured by binary classifiers during the training phase. 
Although such identity information enhances the differences between real and fake images when testing the model on the in-dataset evaluation, it tends to mislead the model on the cross-dataset evaluation.
In this section, we conduct thorough experiments to verify this hypothesis.
First, we conduct the ID linear classification experiment to verify that binary classifiers capture identity information during the training phase. 
Second, we quantified the influence of such ID representation on the in-dataset evaluation and cross-dataset evaluation respectively, to verify its effect on the task of deepfake detection. 
\begin{figure}[t]
\centering
 \begin{subfigure}{0.49\textwidth}
  \includegraphics[width=0.48\textwidth]{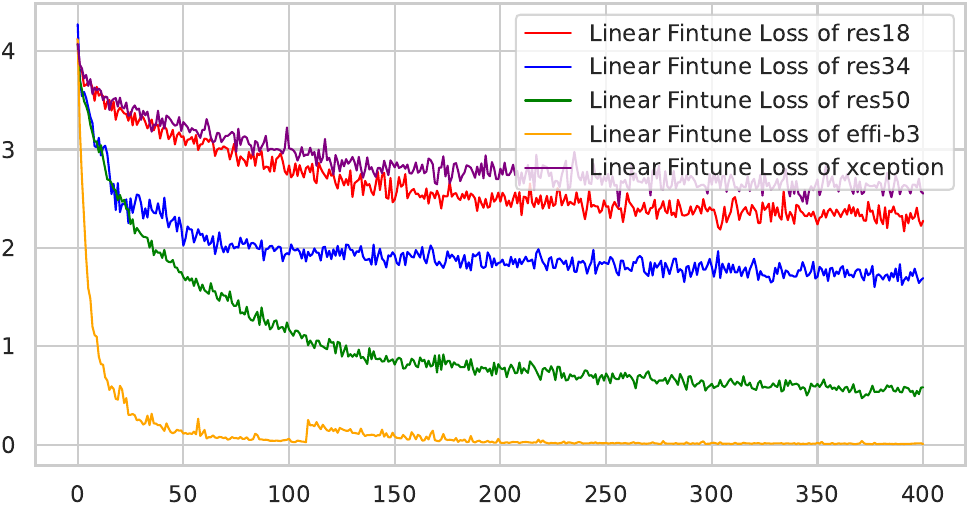}
  \includegraphics[width=0.48\textwidth]{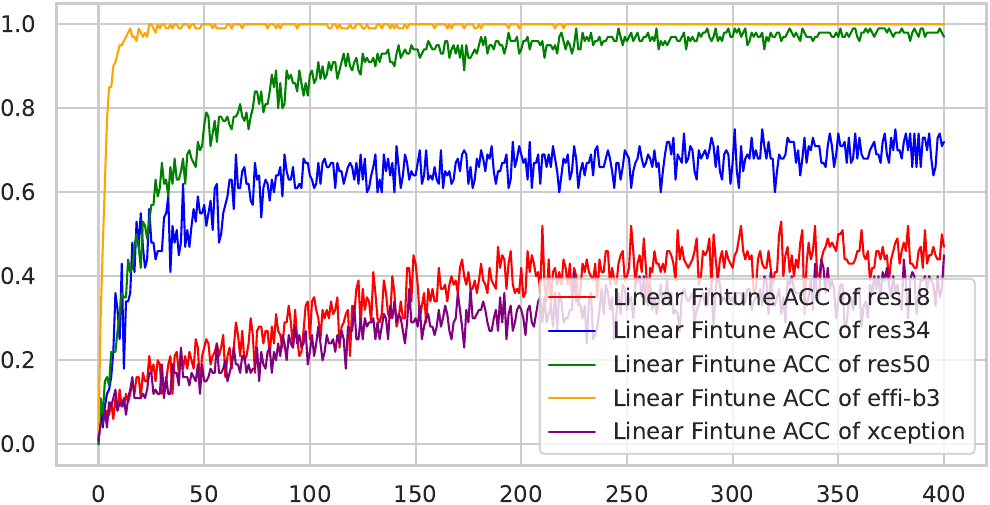}
  \caption{\scriptsize{Celeb-DF}}
  \label{sub:celeb-loss-acc}
  \end{subfigure}
  \begin{subfigure}{0.235\textwidth}
  \includegraphics[width=\textwidth]{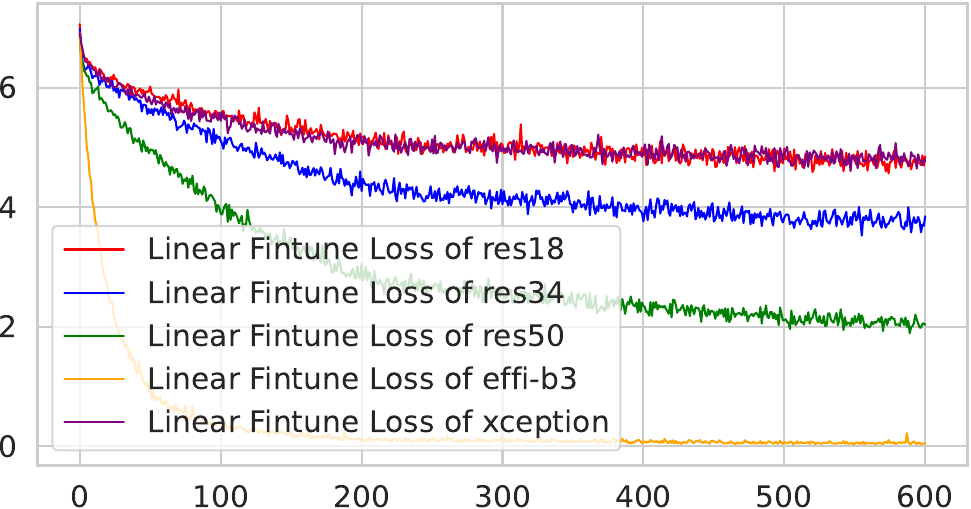}
  \caption{\scriptsize{FF++}}
  \label{sub:b2}
  \end{subfigure}
  \begin{subfigure}{0.235\textwidth}
  \includegraphics[width=\textwidth]{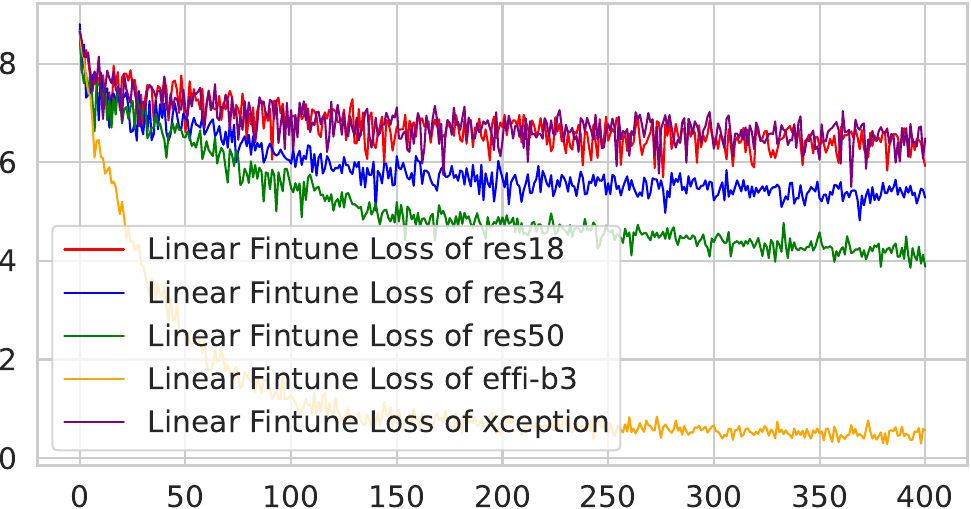}
  \caption{\scriptsize{LFW}}
  \label{sub:b2}
  \end{subfigure}
  
\caption{
\textbf{ID linear classification on frozen features of binary classifiers.}
Results show that binary classifiers of different backbones learned ID representation of images, even without explicit supervision of identity labels.
}
\label{fig:ID_linear_cls}
\end{figure}
\subsection{Verifying the Existence of ID Representation}
\noindent\fbox{
  \parbox{0.45\textwidth}{\textbf{Hypothesis 1}: The ID representation in the deepfake dataset is accidentally captured by binary classifiers during the training phase when without explicit supervision.}
}
~\\
In this section, we performed the ID linear classification experiment to verify that binary classifiers accidentally learn the ID representation on images.

Inspired by previous unsupervised pre-training methods \cite{he2020momentum,chen2020simple}, we finetuned the frozen features extracted from classifiers to evaluate the generalization of the learned ID representation.
Given a binary classifier trained on FF++ \cite{ff++}, we measured the linear classification accuracy of identities on features extracted from the classifier for FF++ \cite{ff++}, Celeb-DF \cite{celeb} and a face recognition dataset LFW \cite{LFW}. 
To be specific, we froze the input feature to the last linear layer of ResNet-18/34/50 \cite{resnet}, Xception \cite{xception} and Efficient-b3 \cite{efficientnet} to demonstrate the universality of such phenomenon. 
Fig. \ref{fig:ID_linear_cls} shows that linear classification on features of different classifiers converged to varying degrees and achieved varying degrees of accuracy for identity classification.
Such results also indicate that although classifiers were never trained on Celeb-DF and LFW before, they still extracted substantial information about identities from images, especially on strong backbones (\emph{e.g.}, Efficient-b3).
In other words, deepfake detectors accidentally learned the ID representation of images, without explicit supervision in particular.
\begin{table}[t]
\begin{center}
{\linespread{1.0}
\setlength\tabcolsep{4pt}
\scriptsize
\begin{threeparttable}
\begin{tabular}{l c c c c c}
\hline
{Datasets}
& ResNet-18 & ResNet-34 & ResNet-50 & Xception & EfficientNet-b3 \\
\hline
\hline
FF++  & 81.53 & 89.77 & 99.58 & 97.32 & 94.87 \\
Celeb-DF & 46.88 & 47.22 & 49.47 & 47.23 & 44.43 \\
\hline
\end{tabular}
\end{threeparttable}
}
\end{center}
\caption{
\textbf{Quantifying the influence of the ID representation on the task of deepfake detection.}
Results show that although ID representation could boost the performance of the in-dataset evaluation, \emph{i.e.} FF++, it would hinder the improvements of the cross-dataset evaluation, \emph{i.e.} Celeb-DF.
}
\label{tb:IIL_quantify}
\end{table}

\subsection{Quantifying the Influence of ID Representation}

\noindent\fbox{
  \parbox{0.45\textwidth}{\textbf{Hypothesis 2}: Although the accidentally learned ID representation may enhance the performance on the in-dataset evaluation, it tends to mislead the model on the cross-dataset evaluation.}
}
~\\
After verifying the existence of ID representation in features of binary classifiers, we performed another experiment to verify its effect on deepfake detection.

The key challenge is how to attribute the output of the binary classifier to the ID representation of the input image quantitatively. 
Intuitively, the identity of an image is not decided by each image region individually, \emph{e.g.} mouths, eyes and noses.
Instead, these regions usually collaborate with each other to form a certain pattern, \emph{e.g.} the identity of the input image.
Thus, we used the multivariate interaction metric \cite{zhang2021interpreting} to quantify the influence of the ID representation.
Such a metric can be considered as the attribution score disentangled from the output score of the input image, which is assigned to the interaction of multiple units.
Let $N=\{1, 2, 3, ..., n\}$ denote all the units of an input image. 
The multivariate interaction caused by the subset of units $S\subseteq N$ is calculated as 
\begin{equation}
    I([S]) = \phi([S]|N_{[S]}) - \sum_{i\in S} \phi(i|N_i)
\end{equation}
where $\phi([S]|N_{[S]})$ denotes the Shapley value \cite{shapley-value} of the coalition $[S]$, which indicates the contribution of $[S]$ to the output score.
$\phi(i|N_i)$ denotes the Shapley value of the unit $i$, which indicates the contribution of the unit $i$. 
$N_{S} = N/S\cup \{[S]\}$ and $N_i = N/i\cup \{i\}$.

In practice, to reduce the computational cost, we sampled $5$ frames from each video and divided the input image into $16\times16$ girds.  
$S$ was set as $S = N$ in experiments, since the input faces are usually cropped and aligned to expand the whole images as a common protocol for the deepfake detection \cite{multi,amazon}.
In this way, we used $I([N])$ as the score of each image and calculated the frame-level AUC to show the effect of ID representation.

\noindent \textbf{Understanding $I([N])$.} $I([N])$ can be rewritten as $I([N]) = \phi([N]|N_{[N]}) - \sum_{i\in N} \phi(i|N_i) = [f(N) - f(\emptyset)] - \sum_{i\in N}[f(\{i\}) - f(\emptyset)]$. 
Here $f$ denotes the binary classifier, which outputs a scalar score for an input image.
Then, we have  $f(N) = I([N]) + \sum_{i\in N}[f(\{i\}) - f(\emptyset)] + f(\emptyset)$.
$f(\emptyset)$ is a constant, representing the output of the classifier when setting the input image as the baseline value.
In this way, we can disentangle the output score of the classifier $f(N)$ into two variables, \emph{i.e.} $I([N])$ and $\sum_{i\in N}[f(\{i\}) - f(\emptyset)]$, which measures the overall interaction among multiple units and the local utility of each unit respectively.
Therefore, when setting $I([N])$ as the output score for each image, if the calculated frame-level AUC $>$ 0.5, $I([N])$ can be considered to contribute positively to the classification task. 
Meanwhile, if the calculated frame-level AUC $<$ 0.5, $I([N])$ can be considered to contribute negatively. 

Results are shown in Table \ref{tb:IIL_quantify}.
We used ResNet-18/34/50, Xception and Efficient-b3 as the backbones of classifiers to clarify the broad effect of ID representation for deepfake detection models.
All classifiers were trained on FF++.
In general, when tested during the in-dataset evaluation, all classifiers achieved AUC $>$ 0.5, indicating the enhancement for differences between real and fake images.
In contrast, all classifiers achieved AUC $<$ 0.5 when tested during the cross-dataset evaluation, which indicates the misguidance of ID representation.
Such results verify our hypothesis.

\begin{figure}[t]
\centering
\includegraphics [width=0.48\textwidth]{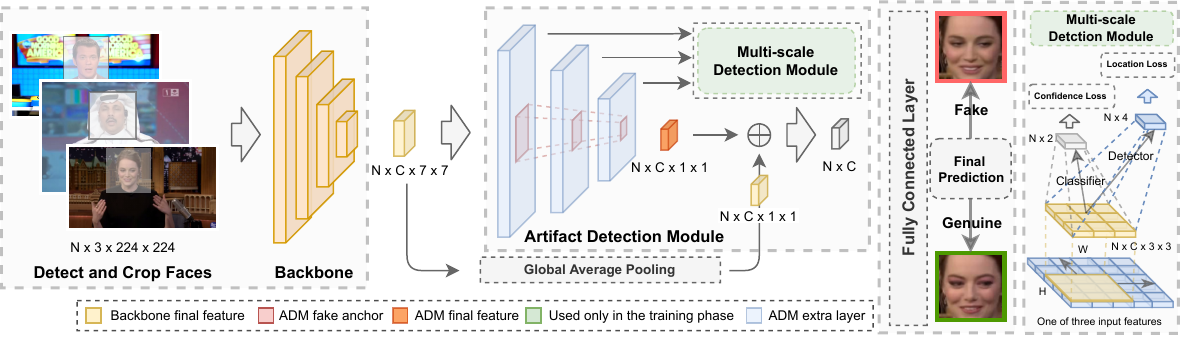}
\caption{
\textbf{The overall framework of the ID-unaware Deepfake Detection Model.} 
$N$ and $C$ denote the number of images and channels.
With the help of the Artifact Detection Module, our model aims to focus on the local representation of images to indicate face forgeries. 
}
\label{fig:CADDM}
\end{figure}

\section{ID-unaware Deepfake Detection Model}
Based on the understanding of the Implicit Identity Leakage, we further propose a simple yet effective method termed as the ID-unaware Deepfake Detection Model to improve the generalization ability of binary classifiers.

\noindent \textbf{Motivation.} 
It has been widely acknowledged that object detection modules focus on local areas of images, instead of the global representation \cite{luo2016understanding,zhang2017s3fd,liu2018receptive,li2019scale}.
Inspired by the fact that local areas usually do not reflect the identity of images, we designed the Artifact Detection Module in our model to focus on local artifact areas on images, so as to pay less attention to the global identity.  
By preventing our model from learning the global ID representation of images, the influence of Implicit Identity Leakage can be reduced.

Moreover, to facilitate the training of the Artifact Detection Module, we propose the Multi-scale Facial Swap method to generate fake images with the ground truth of artifact area positions, which also enriches artifact features in the training phase.
\subsection{Artifact Detection Module}
The overall architecture of the Artifact Detection Module (ADM) is shown in Fig. \ref{fig:CADDM}. 
ADM takes the extracted features from the backbone as the input and detects the position of artifact areas based on multi-scale anchors. 
Specifically, at the end of the backbone, four extra layers of different scales are added, where the sizes of the feature maps decrease following the tuple (7×7, 5×5, 3×3, 1×1).
In the training stage, the Multi-scale Detection Module is placed after the first three extra layers, detecting artifact areas on fake images with multi-scale default anchors on images.
Similar to \cite{ssd,yolov2_anchor,yolov3_anchor}, each feature map grid is associated with multiple default anchors with different scales on the input images. 
The Multi-scale Detection Module adds a detector and a classifier after each extra layer to output the position offsets ($N\times 4$) and confidences of categories ($N\times 2$, \emph{i.e.} the fake or genuine anchor) for each default anchor on images.
A default anchor box is annotated as fake if the Intersection over Union (IoU) between the anchor box and the ground truth of artifact areas is greater than the threshold.
Moreover, the final $1\times 1$ feature maps of ADM create a short connection with the end of the backbone, which further enriches the artifact features learned by the ADM.
Its output is then fed into a fully connected layer to generate the final prediction.

To summarize, ADM determines if there exist artifact areas in multi-scale anchors. 
Such architecture helps our model to pay less attention to the global identity features on images, reducing the influence of Implicit Identity Leakage.

\begin{figure}[t]
\centering
\includegraphics [width=0.45\textwidth]{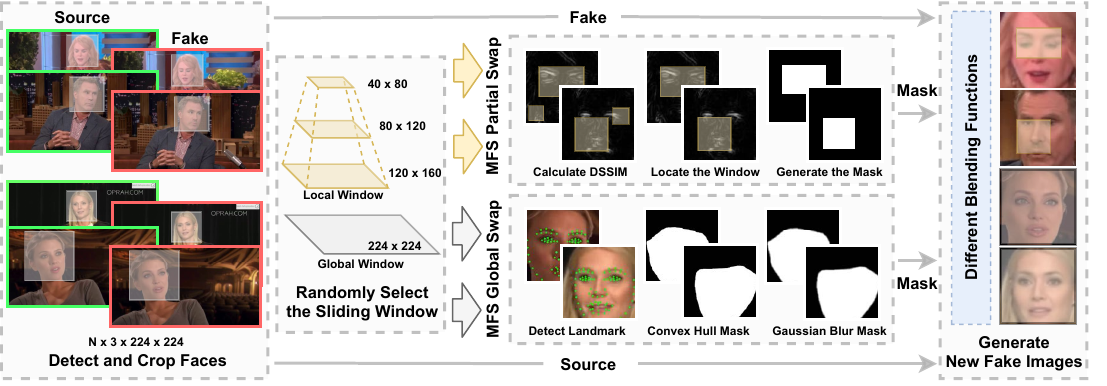}
\caption{
\textbf{Overview of the Multi-scale Facial Swap (MFS).}
MFS manipulates the paired fake image and source image in two ways, namely global swap, and partial swap, to generate new fake images with the ground truth of artifact areas. }
\label{fig:preprocess}
\end{figure}

\subsection{Multi-scale Facial Swap}
The training of the Artifact Detection Module requires fine and local position annotations of artifact areas on images, which are usually not available in public deepfake datasets \cite{ff++,celeb}.
To this end, we propose the Multi-scale Facial Swap (MFS) method, which uses multi-scale sliding windows and different blending functions to create new fake images with the position annotations of artifact areas. 
Besides, the new fake images also further enrich artifact features in the training set.

The procedure of MFS is shown in Fig. \ref{fig:preprocess}.
To generate the new fake image with the position annotations of artifact areas, MFS manipulates the paired
fake image and source image in two ways, i.e. global swap and partial swap.
During the procedure of partial swap, MFS firstly selects a sliding window of a random size to locate the artifact area.
In order to find the local area where artifacts most likely exist, the sliding window is selected by the following equation:
\begin{equation}
  x_t, y_t = \arg\max_{x, y}\sum_{i=x}^{x+h} \sum_{j=y}^{y+w} \text{DSSIM}(I_F, I_S)_{i, j}.
\label{sw_dssim_sum}
\end{equation}
where $DSSIM(\cdot)$ indicates the structural dissimilarity \cite{dssim}, a larger value of which usually suggests that the areas are more probable to contain artifacts.
$x$, $y$ denote the top-left position of the sliding window on images.
$h$, $w$ denote the height and width of the sliding window.
$I_F$, $I_S$ denote the fake image and source image.
Based on the selected sliding window, we then calculate a mask $M$ to generate the new fake image.
Specifically, we get the ground truth of the artifact area by cropping out the sliding window area on the fake image, and generating a new fake image ($I_F ^\prime$) as follows:
\begin{equation}
  I_F^\prime = \text{BLENDING}(I_F, I_S, M).
\label{generate_fake}
\end{equation}
where BLENDING$(\cdot)$ denotes different blending methods (\emph{e.g.}, Poisson blending \cite{poisson} and alpha blending \cite{face-x-ray}).
Take alpha blending \cite{face-x-ray} as an example: $I_F^\prime = I_F * M + I_S * (1-M)$.
The artifact area position of $I_F^\prime$ is [$x_{t}$, $y_{t}$, $x_{t}+h$, $y_{t}+w$]. 
During the procedure of global swap, the sliding window size is equal to the source image. 
As shown in Fig \ref{fig:preprocess}, MFS then generates the new fake image similar to Face-x-ray \cite{face-x-ray}, which also provides more diverse face regions with elastic deformation \cite{ronneberger2015u}.

Overall, with multi-scale sliding windows and different blending methods, MFS generates fake images with ground truth of artifact area positions. 
It supports the training of our model and further enriches artifact features in the dataset.

\subsection{Loss Function}
The overall loss function is a weighted sum of the global classification loss $L_{cls}$ and detection loss $L_{det}$,
\begin{equation}
  L = \beta L_{det} + L_{cls}.
 \label{loss_cadm}
\end{equation}
where $\beta$ is a positive scalar which controls the trade-off between $L_{det}$ and $L_{cls}$. 

$L_{cls}$ is the cross entropy loss to measure the accuracy of the final prediction, \emph{i.e.} fake or genuine images.
$L_{det}$ is the detection loss to guide the learning of ADM.
Similar to \cite{ssd,yolov1_confidence_loc_loss,fast-rcnn}, it contains confidence loss ($L_{conf}$) and location loss ($L_{loc}$).
$L_{conf}$ is the binary cross-entropy loss to measure the predicted result for each anchor, \emph{i.e.} the fake or genuine anchors.
$L_{loc}$ is a Smooth L1 loss \cite{fast-rcnn} to measure the position offsets between ADM predictions and the ground truth of artifact areas,
\begin{equation}
    L_{det} = \frac{1}{N} (L_{conf}(x, c) + \alpha L_{loc}(x, l, g)).
\label{det_loss}
\end{equation}
where $N$ is the number of positive anchor boxes, \emph{i.e.}, fake anchors; $x\in \{0,1\}$ is an indicator for matching the default anchor to the ground truth of artifact areas; $c$ denotes the class confidences; $l$ and $g$ denote the ADM predicted box and the artifact area ground truth box; $\alpha$ denotes a positive weight.
Please see supplementary materials for details.

\section{Experiment}


In this section, we first introduced our experimental settings. 
Then, we compared our model with binary classifiers in terms of the Implicit Identity Leakage, to show the effectiveness of our method.
After that, we explored the contribution of each component in our model.
Finally, we compared our approach with other SOTA deepfake detection methods. 

\subsection{Experiment Setting}
\noindent \textbf{Datesets.}
We trained our models on the widely-used dataset FaceForensics++ (FF++)  \cite{ff++}. 
FF++ contains 4320 videos, \emph{i.e.} 720 original videos collected from YouTube and 3600 fake videos generated by FaceShifter \cite{faceshifter}, FaceSwap \cite{faceswap}, Face2Face \cite{face2face},  Deepfakes \cite{deepfakes} and NeuralTextures \cite{neural-textural}.

We evaluated our approach performance on the following datasets:
(1) FF++ \cite{ff++}, which contains 140 original videos and 700 fake videos.
(2) DFDC-V2 \cite{dfdc}, which has 2500 real videos and 2500 fake videos.
(DFDC-V2 is widely acknowledged as the most challenging data set, since its real videos are close to life, while the artifact areas in its forgery videos are smaller than other datasets).
(3) Celeb-DF \cite{celeb} includes 178 real videos and all fake videos are generated by only one forgery algorithm.



\noindent \textbf{Implementation Details.}
During the training phase, we set the batch size to 128 and the image size to $224\times 224$. MFS sliding window scale was randomly selected from [40, 80], [80, 120], [120, 160], [224, 224].
All images were aligned by face landmarks which were extracted by a detector \cite{mtcnn}. 
Similar to \cite{amazon,face-x-ray}, we also used regular data augmentations (DA) to further improve the model generalization, such as Random Crop, Gaussian Blur/Noise, and JPEG Compression.
We select common classification models pre-trained on ImageNet \cite{imagenet} as the model backbones, including ResNet-34 \cite{resnet} and EfficientNet-b3/b4 \cite{efficientnet}.
We set the number of total epochs to 200, each of which had 512 randomly selected mini-batches.
$\alpha$ and $\beta$ in the loss function were set to 1 and 0.1 by default. 
The learning rate was set to $3.6\times 10^{-4}$ at initialization and decreased to $1\times 10^{-4}$ and $5\times 10^{-5}$ at epoch 10 and epoch 20 respectively for fine-tuning. 
We used Adam \cite{adam} as our optimizer.
In the inference process, we chose 32 frames at an equal interval from each video, using deepfake detection video-level AUC following \cite{ff++,face-x-ray} to report detector performance.

\begin{figure}[t]
\centering
  \begin{subfigure}{0.30\textwidth}
  \includegraphics[width=0.48\textwidth]{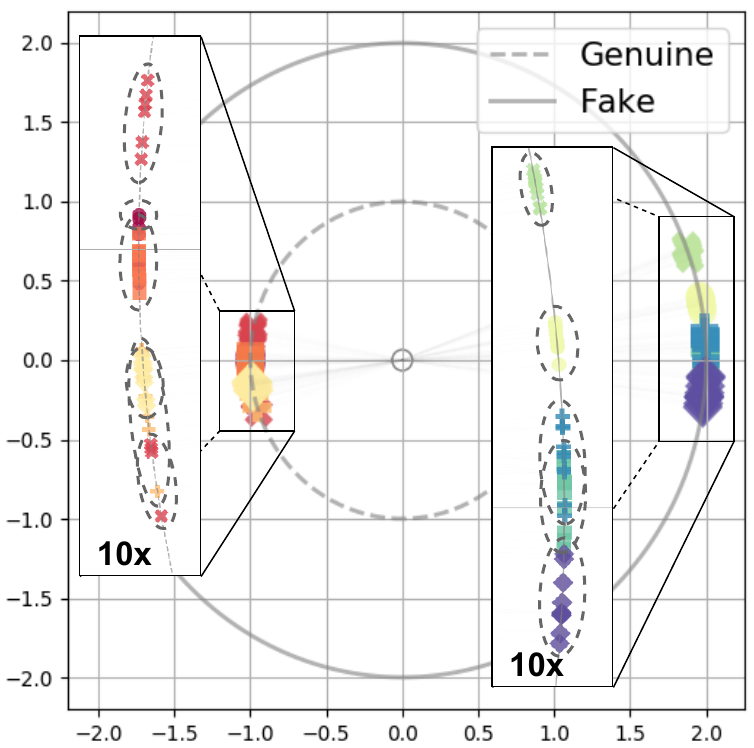}
  \includegraphics[width=0.48\textwidth]{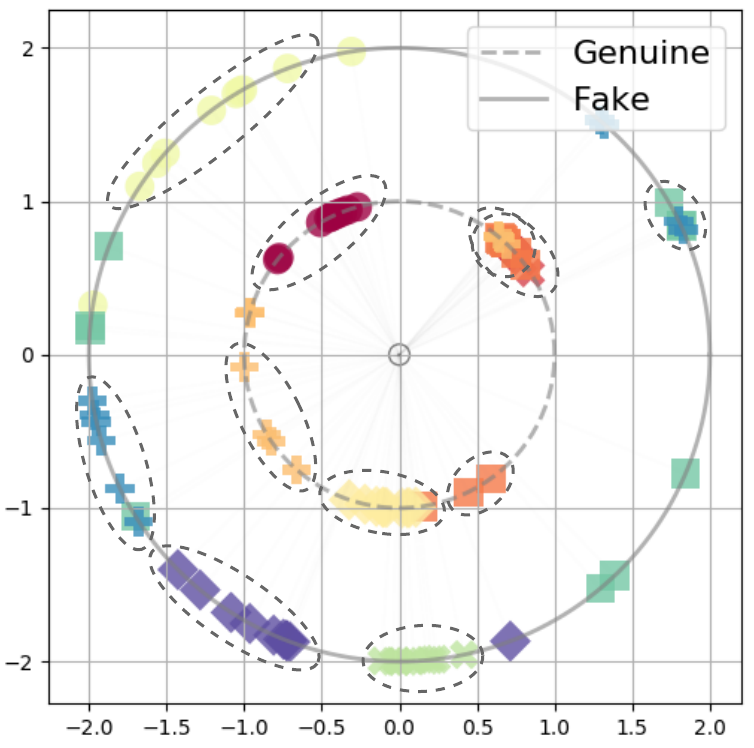}
  \caption{\scriptsize{Binary classifiers (FF++ (left), Celeb-DF (right))}}
  \label{sub:a2}
  \end{subfigure}
  \begin{subfigure}{0.145\textwidth}
  \includegraphics[width=\textwidth]{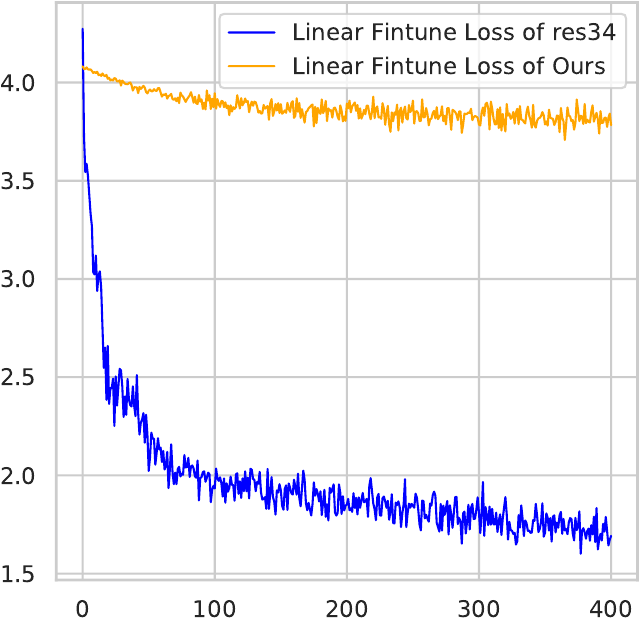}
  \caption{\scriptsize{Training Loss}}
  \label{sub:b2}
  \end{subfigure}
  \begin{subfigure}{0.30\textwidth}
  \includegraphics[width=0.48\textwidth]{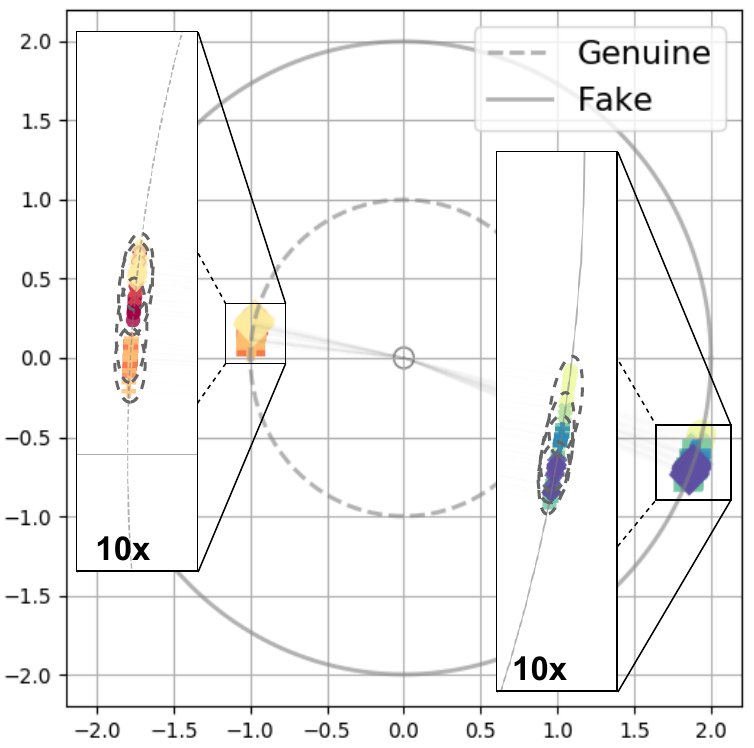}
  \includegraphics[width=0.48\textwidth]{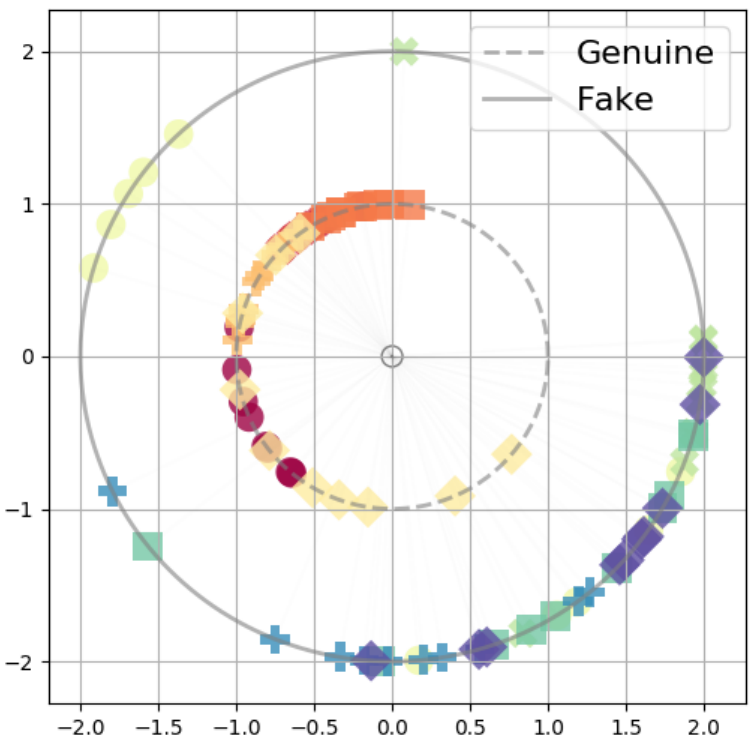}
  \caption{\scriptsize{Ours (FF++ (left), Celeb-DF (right))}}
  \label{sub:c2}
  \end{subfigure}
  \begin{subfigure}{0.145\textwidth}
  \includegraphics[width=\textwidth]{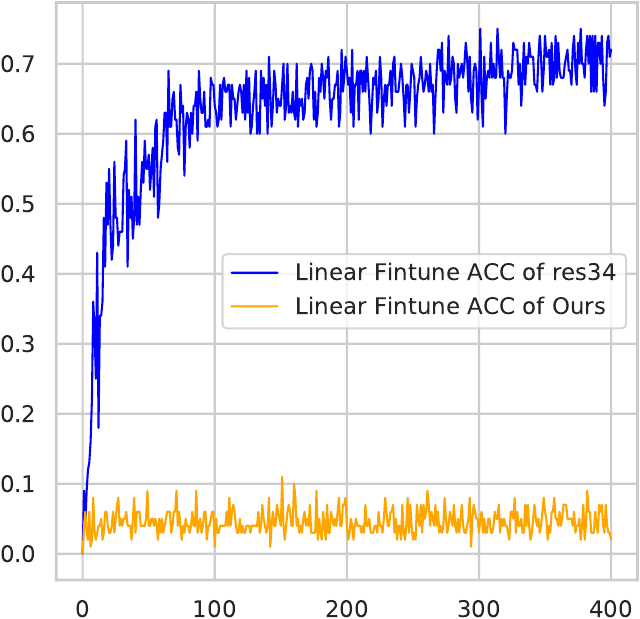}
  \caption{\scriptsize{Training Accuracy}}
  \label{sub:d2}
  \end{subfigure}
\caption{
\textbf{Comparison in terms of the Implicit Identity Leakage between the binary classifier and our model.}
Fig. \ref{sub:a2} and \ref{sub:c2} show the feature space with $L2$ normalization of the binary classifier and our model. 
Fig. \ref{sub:b2} and \ref{sub:d2} show the linear classification of identities on features of our model and the binary classifier.
Results show that our model paid less attention to the ID representation of images.
}
\label{fig:relation}
\end{figure}

\subsection{Experimental Analysis}
\label{sec:IIL2}
\noindent \textbf{Comparison of Implicit Identity Leakage.}
We designed experiments to compare our model with the binary classifier in terms of the phenomenon of Implicit Identity Leakage, in order to demonstrate that our model alleviates this problem.
Each model used the EfficientNet-b3 as the common backbone and was trained on the FF++ dataset.
We randomly sampled 100 images with 10 identities and used t-SNE \cite{van2008visualizing} to visualize the high dimensional features extracted from the final layer of different models in 2D.
Each point represents the features of an image. 
Different markers of points represent features of images with different identities.

In Fig. \ref{sub:a2} (left), on the in-dataset evaluation, the binary classifier successfully distinguished the differences between the fake images and genuine images. 
However, features of different identities were visually separable, which shows the binary classifier partially learned the identity information of images.
When tested in the Celeb-DF, as shown in Fig. \ref{sub:a2} (right), such unnecessary knowledge about identity information tended to be misused by the binary classifier, hindering its performance.

In contrast,  on the in-dataset evaluation for our model (Fig. \ref{sub:c2} (left)), features of different identities were visually inseparable and overlapped with each other, which shows that our model reduced the influence of Implicit Identity Leakage. 
When tested in the Celeb-DF (Fig. \ref{sub:c2} (right)), our model indicated fake images by detecting artifact areas with less influence of the identity information and still roughly distinguished the differences between fake images and genuine images.
Such results show that ADM helped our model alleviate the phenomenon of Implicit Identity Leakage. 
Moreover, we also conduct further experiments to quantitatively compare our model with the binary classifier in terms of the phenomenon of Implicit Identity Leakage. 
Please see supplementary materials for more analysis.

Meanwhile, we conduct the same ID linear classification experiment as before to compare the existence of ID representation in features of our model and the binary classifier.
We used ResNet-34 as the backbone.
Fig. \ref{sub:b2} and \ref{sub:d2} show that linear classification on features of binary classifiers was easier to converge and achieved better accuracy than features of our model.
Such results indicate that features of binary classifiers contained more information about identities than our model, further verifying the phenomenon of Implicit Identity Leakage (IIL) and the remedy effect of our method to alleviate its negative influence.  

\subsection{Ablation Studies}
\begin{table}[t]
\begin{center}
{\linespread{1.0}
\setlength\tabcolsep{4pt}
\scriptsize
\begin{threeparttable}
\begin{tabular}{l c c c c c c}

\hline
\multirow{2}{*}{Models}
&\multirow{2}{*}{DA} &\multirow{2}{*}{MFS} &\multirow{2}{*}{ADM} 
&\multicolumn{3}{c}{Test Set (AUC (\%)) } \\
\cline{5-7}

~ & ~ & ~ & ~ & FF++ & Celeb-DF  & DFDC-V2 \\

\hline
\hline

\multirow{5}{*}{ResNet-34} 
& $\times$ & $\times$ & $\times$  & \textbf{99.88} & 64.05  & 48.73 \\

& $\times$ & \checkmark & $\times$ & 98.70 & 76.35  ($\uparrow${12.30})  & 59.97  ($\uparrow${11.24}) \\

& \checkmark & $\times$ & $\times$ & 99.74 & 80.07 ($\uparrow${16.02}) & 62.46 ($\uparrow${13.73})\\

& \checkmark & \checkmark & $\times$ & 99.75 & 86.68  ($\uparrow${22.63})  & 67.94  ($\uparrow${19.21}) \\

& \checkmark & \checkmark & \checkmark  & 99.70 & \textbf{91.15}  ($\uparrow${27.10}) & \textbf{71.49}  ($\uparrow${22.76}) \\

\hline
\end{tabular}
\end{threeparttable}

}
\end{center}
\caption{
\textbf{Experimental results for the effect of different components of our model.}
Here DA denotes the Data Augmentations. 
Each model was trained by FF++ and tested on FF++, Celeb-DF, and DFDC-V2. 
Our model shows a significant improvement in cross-dataset evaluation.
}
\label{tb:eff-caddm}
\end{table}
\noindent \textbf{Effect of Different Components.}
To confirm the effectiveness of our model, we evaluated how data augmentations (DA), MFS, and ADM affected the accuracy of our model.
We trained models on FF++  and tested the performance on FF++, Celeb-DF, and DFDC-V2. 
We denote the model without DA, MFS, and ADM as the baseline. 
As shown in Table \ref{tb:eff-caddm}, the baseline achieved the best performance in the in-dataset evaluation.
However, the baseline only achieved $64.05$\% and $48.73$\% of AUC on Celeb-DF and DFDC-V2.

Similar to \cite{amazon,face-x-ray}, we also added regular data augmentations (DA), such as Random Crop, Gaussian Blur/Noise, and JPEG Compression to the baseline. 
As shown in Table \ref{tb:eff-caddm}, when DA were added, our model got AUC improvements of $16.02$\% and $13.73$\% on Celeb-DF and DFDC-V2 compared with the baseline.
We argue that these augmentations may disrupt the identity information of the data to some degree and improve the model generalization. 

Besides, ADM and MFS further improved the cross-dataset evaluation performance.
ADM guided our model to learn artifact representations of local areas, thus reducing the influence of Implicit Identity Leakage.
MFS-generated images shared similar identity information with source images (see Fig. \ref{fig:visual} for example).
Such aligned ID representation eliminated the misguidance of ID information between genuine and fake images to some degree, which also helped to reduce the influence of Implicit Identity Leakage. 
As shown in Table \ref{tb:eff-caddm}, beyond the effect of DA, our proposed method (\emph{i.e.}, MFS and ADM) still led to \emph{further yet remarkable} improvements (\emph{e.g.},  11.08\% on Celeb-DF and 9.03\% on DFDC-V2 in Tab.2).
Such improvements based on a strong baseline model (\emph{i.e.}, a model trained with DA.) could further validate the efficacy of our proposed method \emph{w.r.t.} a weak baseline model.

\begin{table}[t]
\begin{center}
{\linespread{1.0}
\setlength\tabcolsep{8 pt}
\scriptsize
\begin{threeparttable}
\begin{tabular}{l | c | c  c}

\hline

\multirow{2}{*}{Models}
&\multirow{2}{*}{Backbones}
&\multicolumn{2}{c}{Test Set (AUC (\%)) } \\
\cline{3-4}
~ & ~ & FF++ & Celeb-DF\\

\hline
\hline

Multi-task \cite{nguyen2019multi} & - & 76.30 & 54.30 \\

Xception \cite{ff++} & Xception  &99.58 &49.03 \\

MMMS \cite{wang2021m2tr} & Transformer  & 99.50 & 65.70 \\

SPSL \cite{spatial} & Xception 
&96.91 &76.88 \\

Local-Relation \cite{chen2021local} & -  & - & 78.26 \\

Two-branch \cite{masi2020two} & DenseNet  & 93.20 & 73.40 \\

\textbf{}DSP-FWA \cite{li2018exposing}
& ResNet-50  &93.00 &64.60 \\

$F^3$-Net \cite{qian2020thinking}
& Xception  & 98.10 & 65.17 \\

MAT \cite{multi} & Efficient-b4 
&99.61 &68.44 \\

SLADD \cite{chen2022self} & Xception
&98.40 & 79.70\\

Face-x-ray \cite{face-x-ray} & HRNet 
&99.17 &80.58 \\

PCL+I2G \cite{amazon} & ResNet-34 
&99.11 &90.03 \\

SBI \cite{SBI} & Efficient-b4
&99.64 &93.18 \\
\hline
\hline

\multirow{3}{*}{Ours} & ResNet-34
&\textbf{99.70}($\uparrow${0.06}) 
&91.15 \\

& Efficient-b3 
&\textbf{99.78}($\uparrow${0.14}) 
&93.08 \\

& Efficient-b4 
&\textbf{99.79}($\uparrow${0.15}) 
&\textbf{93.88}($\uparrow${0.70})  \\
\hline

\end{tabular}
\end{threeparttable}

\linespread{1.0}
\setlength\tabcolsep{1pt}
\scriptsize
\begin{threeparttable}
\begin{tabular}{l | c c c c c | c c c}
\hline
\multirow{2}{*}{Datasets}
&\multirow{2}{*}{Xception}
&\multirow{2}{*}{SPSL}
&\multirow{2}{*}{PCL+I2G}
&\multirow{2}{*}{MAT}
&\multirow{2}{*}{SBI}
&\multicolumn{3}{c}{Ours} \\[1ex] \cline{7-9}
~ &\cite{ff++} &\cite{spatial} &\cite{amazon} &\cite{multi} &\cite{SBI} & {Res-34} & {Effi-b3} & Effi-b4 \\

\hline
\hline

DFDC-V2 & 45.60 & 66.16 & 67.52 & 70.99 &72.42
& 71.49 
& \textbf{73.74} &\textbf{73.85}($\uparrow$1.43)\\

\hline

\end{tabular}
\end{threeparttable}
}
\end{center}
\caption{
\textbf{Comparison with the SOTA in FF++ (top), Celeb-DF (top) and DFDC-V2 (bottom)}. All methods were trained on FF++.
Some numbers are missing because these methods do not provide training codes or pre-trained models. 
}

\label{tb:compare-sota}

\end{table}

\subsection{Comparison with state-of-the-art methods}
\begin{figure}[t]
\centering
\includegraphics [width=0.48\textwidth]{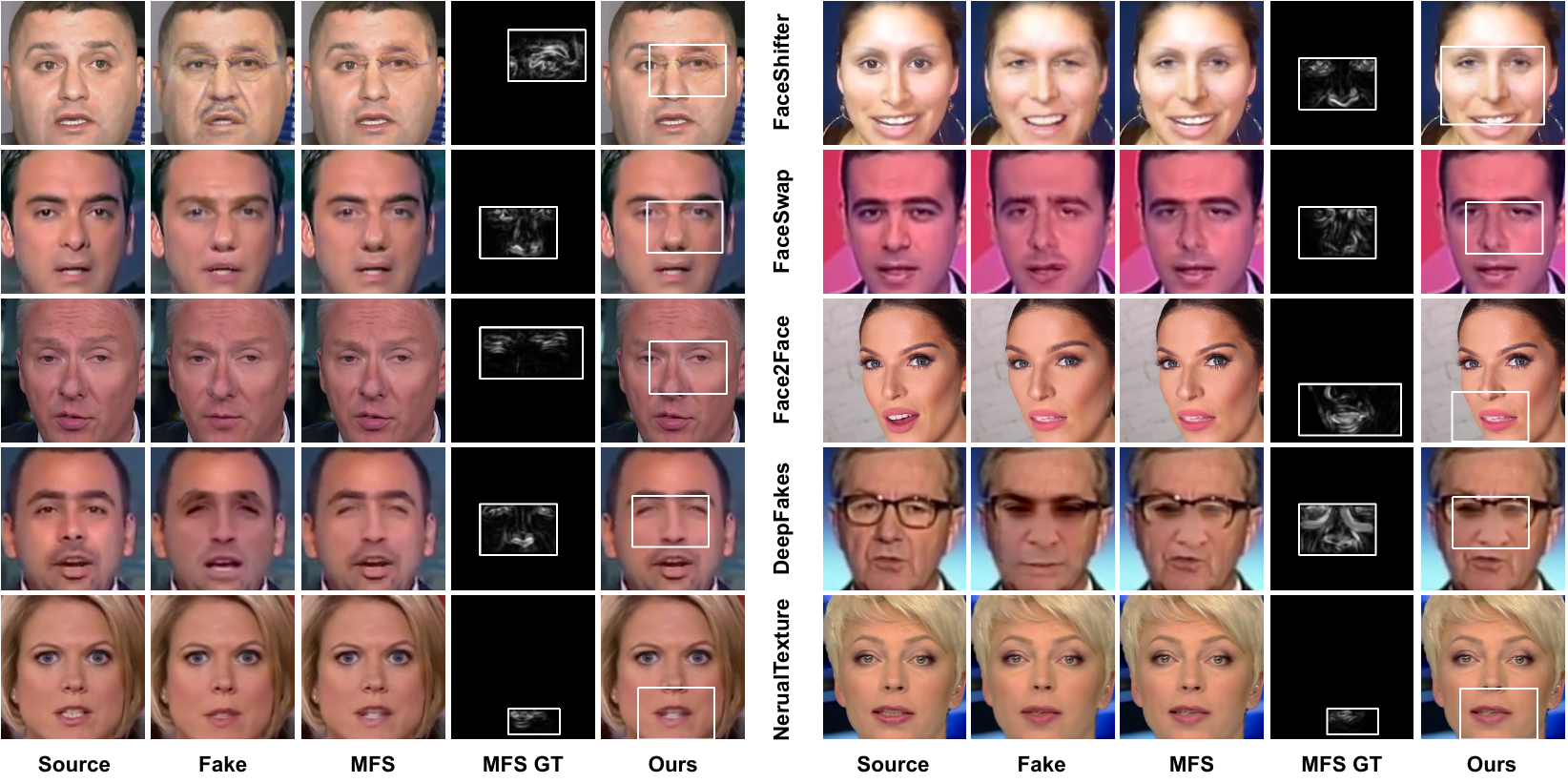}
\caption{
\textbf{Visual results on various facial manipulation algorithms.}
We used NMS in \cite{ssd,yolov1_confidence_loc_loss,fast-rcnn} to select the bounding box position with the highest score as the final prediction. When the score was less than the threshold, no anchor box would be predicted, as was the case for real images.
The result shows that our model indicated fake images based on local artifact areas.
}
\label{fig:visual}
\end{figure}

As shown in Tab. \ref{tb:compare-sota}, we compared our model with other deepfake detection methods on three public deepfake detection datasets. 
In Tab. \ref{tb:compare-sota}, models were trained on FF++ and tested on FF++, Celeb-DF and DFDC-V2.
Moreover, we also provided visual results of our method in Fig. \ref{fig:visual}, which demonstrate that our model successfully indicates fake images manipulated by different deepfake algorithms and regresses the bounding box of the artifact areas.







\noindent \textbf{In-dataset Evaluation.}
Previous methods exploiting binary classifiers usually achieved great performances on the in-dataset evaluation (\emph{e.g.} $99.58$\% on FF++ for Xception \cite{ff++}.).
Meanwhile, hand-crafted methods forced models to learn specific artifact features on images, which limited model performance on the in-dataset evaluation a bit (\emph{e.g.} $99.17$\% on FF++ for Face-x-ray \cite{face-x-ray}).
Compared with the above methods, by reducing the influence of Implicit Identity Leakage, our method automatically learned various artifact features on images and achieved even better performance on the in-dataset evaluations.
Specifically, compared with the best performing method SBI \cite{SBI}, our approach (Efficient-b4 based) still improved AUC $0.15$\% on the FF++ for the in-dataset evaluation.

\noindent \textbf{Cross-dataset Evaluation.}
Hand-crafted methods forced models to learn specific artifact features on images, improving the performance on the cross-dataset evaluation (\emph{e.g.} $80.58$\% on Celeb-DF for Face-x-ray \cite{face-x-ray}). 
These methods can also be seen as reducing the influence of IIL by guiding models to learn hand-crafted artifact features, instead of ID-relevant information.
Nevertheless, such human-defined artifact features may fail to reflect the generalized representations inside all manipulated areas, which limits their further improvements.
In contrast, by reducing the influence of Implicit Identity Leakage, our method learned generalized artifact features on forgery data, achieving better generalization in the cross-dataset evaluations.
In particular, in Table \ref{tb:compare-sota} (bottom), our model achieved $1.43$\% higher AUC than SBI \cite{SBI} in the recently released DFDC-V2, which is widely considered as the most challenging dataset.
Moreover, our approach also achieved $0.70\%$ higher AUC for the cross-dataset evaluation on Celeb-DF in Table \ref{tb:compare-sota} (top).

In summary, compared with previous methods, our method significantly improved the performance on both the in-dataset and cross-dataset evaluations, showing the effectiveness of reducing the influence of Implicit Identity Leakage to learn generalized artifact features on face forgeries.







\begin{table}[t]
\begin{center}
{
\linespread{1.0}
\setlength\tabcolsep{3.8 pt}
\scriptsize
\begin{threeparttable}
\begin{tabular}{l |c c c c c c | c}

\hline

Method & Saturation & Contrast & Block & Noise & Blur & Pixel & \textbf{Avg} \\

\hline
\hline

Xception \cite{ff++} & 99.3&98.6&99.7&53.8&60.2&74.2&81.0\\

Face-x-ray \cite{face-x-ray} & 97.6&88.5&99.1&49.8&63.8& 88.6& 81.2\\

LipForensices \cite{haliassos2021lips} & \textbf{99.9}&99.6&87.4&73.8&96.1& 95.6& 92.1\\

Ours & 99.6&\textbf{99.8}&\textbf{99.8}&\textbf{87.4}&\textbf{99.0}&\textbf{98.8}&\textbf{97.4}\\
\hline
\end{tabular}
\end{threeparttable}
}
\end{center}
\caption{\textbf{Robustness evaluation on FF++}. Results show that our method achieved better robustness image perturbations.}
\label{tb:robustness}
\end{table}

\noindent \textbf{Robustness Evaluation.}
We also evaluated the robustness of our model to different kinds of image perturbations, following the same robustness experiment setting in LipForensics \cite{haliassos2021lips}. 
Results in Tab. \ref{tb:robustness}  show that our method achieved 97.4\% of video-level AUC on average, higher than LipForensics (\emph{i.e.}, 92.1\%).

\begin{table}[t]
\begin{center}
{
\linespread{1.0}
\setlength\tabcolsep{3 pt}
\scriptsize
\begin{threeparttable}
\begin{tabular}{c| c |c c c  c| c}

\hline

Training set & Model & DF & F2F & FS & NT & FF++ \\

\hline
\hline

\multirow{2}{*}{DF} & Xception \cite{ff++} & 99.38 & 75.05 & 49.13 & 80.39 & 76.34\\

& Ours+Xception  \cite{ff++} & \textbf{100.00}	&\textbf{83.94}	&\textbf{58.33}	&\textbf{68.98}&\textbf{77.81}  ($\uparrow${1.47}) \\
\hline

\multirow{2}{*}{F2F}& Xception  \cite{ff++} & 87.56&99.53 &65.23& 65.90 &79.55\\

& Ours+Xception  \cite{ff++} & \textbf{99.88}&	\textbf{99.97}&	\textbf{79.40}&	\textbf{82.38}&\textbf{90.41} ($\uparrow${10.86})\\

\hline

\multirow{2}{*}{FS} & Xception  \cite{ff++} & 70.12& 61.70& 99.36 &\textbf{68.71} &74.91 \\

& Ours+Xception  \cite{ff++} & \textbf{93.42}&	\textbf{74.00}&	\textbf{99.92}&	49.86&\textbf{79.30} ($\uparrow${4.39})\\

\hline

\multirow{2}{*}{NT} & Xception  \cite{ff++} & 93.09& 84.82& 47.98& 99.50& 83.42\\

& Ours+Xception  \cite{ff++} &\textbf{100.00}&	\textbf{97.93}&	\textbf{86.76}&	\textbf{99.46}&\textbf{96.04} ($\uparrow${12.62})\\
\hline
\end{tabular}
\end{threeparttable}
\linespread{1.0}
\setlength\tabcolsep{3.5 pt}
\tiny
\begin{center}
\begin{threeparttable}
\begin{tabular}{c| c | c |c | c | c |c| c}
\hline
\multirow{2}{*}{Training set}
&\multicolumn{1}{c|}{\multirow{2}{*}{Model}}
 &\multicolumn{2}{c|}{Test Set} & \multirow{2}{*}{Training set}
&\multicolumn{1}{c|}{\multirow{2}{*}{Model}}
 &\multicolumn{2}{c}{Test Set} 
 \\ \cline{3-4} \cline{7-8}
&~&\multicolumn{1}{c|}{FF++} &\multicolumn{1}{c|}{DFDC-V2}&~ &~&\multicolumn{1}{c|}{DFDC-V2}&\multicolumn{1}{c}{FF++} \\
\hline
\hline
\multirow{4}{*}{FF++}
& ResNet-34& \textbf{99.88} & 48.73 & \multirow{4}{*}{DFDC-V2}
& ResNet-34& 92.49 & 60.56 \\
& Ours+ResNet-34 & 99.70 & \textbf{71.49} &~& Ours+ResNet-34& \textbf{94.85}& \textbf{77.32}  \\
\cline{2-4} \cline{6-8}

& Effi-b3& 99.75 & 54,12 &~ & Effi-b3& 94.31& 60.87  \\
& Ours+Effi-b3& \textbf{99.78} & \textbf{73.74}&~
& Ours+Effi-b3& \textbf{95.67}& \textbf{84.43} \\
\hline

\end{tabular}
\end{threeparttable}
\end{center}
}
\end{center}
\caption{\textbf{Cross-method evaluation in FF++ (top) and generalization evaluations between FF++ and DFDC-V2 (bottom).} Our method achieved better generalization capabilities.}
\label{tb:crossmethod}
\end{table}




\noindent \textbf{Cross-method Evaluation.} 
We also conducted the cross-method experiment on FF++, where our model was trained on one type of manipulated data and tested on the remaining three. 
Results in Tab. \ref{tb:crossmethod} (top) show that our approach demonstrated superior cross-method generalization to binary classifiers \cite{ff++} under the same setting.
Such results indicate that \textit{the impact of the learned ID representations also hindered the cross-method generalization of binary classifiers for deepfake detection to varying degrees}, which complements the proposed hypotheses in Sec. \ref{sec:IIL}.
Note that compared with previous hand-crafted methods like \cite{face-x-ray}, which additionally enriched the training data with specific artifact features, there still exists a performance gap.

To point out, one of the most important motivations of our method lies in the goal to \textit{train a robust deepfake detection model, which can automatically learn various yet faithful artifact features from the ever-increasing number of training data}.
To this end, by reducing the impact of IIL, Sec. B in supplementary materials shows that our method successfully captured generalized artifact features as the types of training forgeries increased.
Compared with previous hand-crafted methods \cite{face-x-ray,amazon,improving} designed for specific artifact features, such a data-driven training scheme helps to \textit{release the pressure to continuously devise new methods for the newly-proposed types of forgeries}, which is of great value for real-life applications.
Therefore, results in Tab. \ref{tb:crossmethod} (top) are expected due to the limited artifact features in the original single type of forgeries, which, however, could be easily solved by scaling up the training data.

Furthermore, we evaluated the generalization capabilities of our method in a more challenging scenario.
FF++ and DFDC-V2 were collected with both different manipulation methods and different original videos, making the generalization between these datasets even more difficult.
To this end, Tab. \ref{tb:crossmethod} (bottom) demonstrates the efficacy of our method compared with binary classifiers \cite{resnet,efficientnet}.


\begin{table}[t]
\begin{center}
{\linespread{1.0}
\setlength\tabcolsep{9pt}
\scriptsize
\begin{threeparttable}
\begin{tabular}{l | c c c }
\hline
{Models}
&{FF++}
&{Celeb-DF}
&{DFDC-V2}\\
\hline
\hline
SBI \cite{SBI} & \textbf{99.64} & 93.18 & 72.42 \\
Ours+SBI \cite{SBI} & 99.33($\downarrow$ {0.31}) & \textbf{94.15}($\uparrow$ {0.97})  & \textbf{79.57}($\uparrow$ {7.15}) \\
\hline
\end{tabular}
\end{threeparttable}
}
\end{center}
\caption{
\textbf{Combination with the method of SBI \cite{SBI}}. 
Results show that when combining SBI with our method, the performance on the cross-dataset evaluation was significantly improved.}
\label{tb:combine_SBI}
\end{table}

\noindent \textbf{Potential Applicability.}
Since the Artifact Detection Module is convenient to add to the end of different backbones, our method can also be potentially applied to other SOTA methods as a plug-and-play module to further boost performance. 
To simplify the story, we combined our method with SBI \cite{SBI} for instance. 
The model of \cite{SBI} is a binary classifier, which can be easily modified by adding the Artifact Detection Module.
As shown in Table \ref{tb:combine_SBI}, when combined with SBI, our method achieved significantly better results during the cross-dataset evaluation. 
Such results show the potential applicability of our study, which we believe could help advance the field.

\section{Conclusion}
In this paper, we discover the phenomenon termed as Implicit Identity Leakage through experimental verification: the deepfake detection model is sensitive to the identity information of the data, which reduces the model generalization ability on unseen datasets.
To this end, we propose ID-unaware Deepfake Detection Model to alleviate the Implicit Identity Leakage phenomenon.
Extensive experiments demonstrate that by reducing the influence of Implicit Identity Leakage, our model successfully learns generalized artifact features and outperforms the state-of-the-art methods. 
In summary, this research provides a new perspective to understand the generalization of deepfake detection models, which sheds new light on the development of the field.

{\small
\bibliographystyle{ieee_fullname}
\bibliography{IIL}

\begin{thebibliography}{10}\itemsep=-1pt

\bibitem{mesonet}
Darius Afchar, Vincent Nozick, Junichi Yamagishi, and Isao Echizen.
\newblock Mesonet: a compact facial video forgery detection network.
\newblock In {\em 2018 IEEE International Workshop on Information Forensics and
  Security (WIFS)}, pages 1--7. IEEE, 2018.

\bibitem{Alpher02}
FirstName Alpher.
\newblock Frobnication.
\newblock {\em IEEE TPAMI}, 12(1):234--778, 2002.

\bibitem{Alpher03}
FirstName Alpher and FirstName Fotheringham-Smythe.
\newblock Frobnication revisited.
\newblock {\em Journal of Foo}, 13(1):234--778, 2003.

\bibitem{Alpher04}
FirstName Alpher, FirstName Fotheringham-Smythe, and FirstName Gamow.
\newblock Can a machine frobnicate?
\newblock {\em Journal of Foo}, 14(1):234--778, 2004.

\bibitem{Alpher05}
FirstName Alpher and FirstName Gamow.
\newblock Can a computer frobnicate?
\newblock In {\em CVPR}, pages 234--778, 2005.

\bibitem{bayar2016deep}
Belhassen Bayar and Matthew~C Stamm.
\newblock A deep learning approach to universal image manipulation detection
  using a new convolutional layer.
\newblock In {\em Proceedings of the 4th ACM workshop on information hiding and
  multimedia security}, pages 5--10, 2016.

\bibitem{bondi2020training}
Luca Bondi, Edoardo~Daniele Cannas, Paolo Bestagini, and Stefano Tubaro.
\newblock Training strategies and data augmentations in cnn-based deepfake
  video detection.
\newblock In {\em 2020 IEEE International Workshop on Information Forensics and
  Security (WIFS)}, pages 1--6. IEEE, 2020.

\bibitem{chen2022self}
Liang Chen, Yong Zhang, Yibing Song, Lingqiao Liu, and Jue Wang.
\newblock Self-supervised learning of adversarial example: Towards good
  generalizations for deepfake detection.
\newblock In {\em Proceedings of the IEEE/CVF Conference on Computer Vision and
  Pattern Recognition}, pages 18710--18719, 2022.

\bibitem{simswap}
Renwang Chen, Xuanhong Chen, Bingbing Ni, and Yanhao Ge.
\newblock Simswap: An efficient framework for high fidelity face swapping.
\newblock In {\em Proceedings of the 28th ACM International Conference on
  Multimedia}, pages 2003--2011, 2020.

\bibitem{chen2021local}
Shen Chen, Taiping Yao, Yang Chen, Shouhong Ding, Jilin Li, and Rongrong Ji.
\newblock Local relation learning for face forgery detection.
\newblock In {\em Proceedings of the AAAI Conference on Artificial
  Intelligence}, volume~35, pages 1081--1088, 2021.

\bibitem{chen2020simple}
Ting Chen, Simon Kornblith, Mohammad Norouzi, and Geoffrey Hinton.
\newblock A simple framework for contrastive learning of visual
  representations.
\newblock In {\em International conference on machine learning}, pages
  1597--1607. PMLR, 2020.

\bibitem{stargan}
Yunjey Choi, Youngjung Uh, Jaejun Yoo, and Jung-Woo Ha.
\newblock Stargan v2: Diverse image synthesis for multiple domains.
\newblock In {\em Proceedings of the IEEE/CVF Conference on Computer Vision and
  Pattern Recognition}, pages 8188--8197, 2020.

\bibitem{xception}
Fran{\c{c}}ois Chollet.
\newblock Xception: Deep learning with depthwise separable convolutions.
\newblock In {\em Proceedings of the IEEE conference on computer vision and
  pattern recognition}, pages 1251--1258, 2017.

\bibitem{cozzolino2017recasting}
Davide Cozzolino, Giovanni Poggi, and Luisa Verdoliva.
\newblock Recasting residual-based local descriptors as convolutional neural
  networks: an application to image forgery detection.
\newblock In {\em Proceedings of the 5th ACM Workshop on Information Hiding and
  Multimedia Security}, pages 159--164, 2017.

\bibitem{cozzolino2021id}
Davide Cozzolino, Andreas R{\"o}ssler, Justus Thies, Matthias Nie{\ss}ner, and
  Luisa Verdoliva.
\newblock Id-reveal: Identity-aware deepfake video detection.
\newblock In {\em Proceedings of the IEEE/CVF International Conference on
  Computer Vision}, pages 15108--15117, 2021.

\bibitem{imagenet}
Jia Deng, Wei Dong, Richard Socher, Li-Jia Li, Kai Li, and Li Fei-Fei.
\newblock Imagenet: A large-scale hierarchical image database.
\newblock In {\em 2009 IEEE conference on computer vision and pattern
  recognition}, pages 248--255. Ieee, 2009.

\bibitem{ding2020swapped}
Xinyi Ding, Zohreh Raziei, Eric~C Larson, Eli~V Olinick, Paul Krueger, and
  Michael Hahsler.
\newblock Swapped face detection using deep learning and subjective assessment.
\newblock {\em EURASIP Journal on Information Security}, 2020:1--12, 2020.

\bibitem{dfdc}
Brian Dolhansky, Joanna Bitton, Ben Pflaum, Jikuo Lu, Russ Howes, Menglin Wang,
  and Cristian Canton~Ferrer.
\newblock The deepfake detection challenge dataset.
\newblock {\em arXiv e-prints}, pages arXiv--2006, 2020.

\bibitem{dong2022explaining}
Shichao Dong, Jin Wang, Jiajun Liang, Haoqiang Fan, and Renhe Ji.
\newblock Explaining deepfake detection by analysing image matching.
\newblock {\em arXiv preprint arXiv:2207.09679}, 2022.

\bibitem{dong2022protecting}
Xiaoyi Dong, Jianmin Bao, Dongdong Chen, Ting Zhang, Weiming Zhang, Nenghai Yu,
  Dong Chen, Fang Wen, and Baining Guo.
\newblock Protecting celebrities from deepfake with identity consistency
  transformer.
\newblock In {\em Proceedings of the IEEE/CVF Conference on Computer Vision and
  Pattern Recognition}, pages 9468--9478, 2022.

\bibitem{du2019towards}
Mengnan Du, Shiva Pentyala, Yuening Li, and Xia Hu.
\newblock Towards generalizable forgery detection with locality-aware
  autoencoder.
\newblock {\em arXiv e-prints}, pages arXiv--1909, 2019.

\bibitem{durall2019unmasking}
Ricard Durall, Margret Keuper, Franz-Josef Pfreundt, and Janis Keuper.
\newblock Unmasking deepfakes with simple features.
\newblock {\em arXiv preprint arXiv:1911.00686}, 2019.

\bibitem{deepfakes}
FaceSwapDevs.
\newblock {Deepfakes}.
\newblock \url{https://github.com/deepfakes/faceswap}, 2019.

\bibitem{fast-rcnn}
Ross Girshick.
\newblock Fast r-cnn.
\newblock In {\em Proceedings of the IEEE international conference on computer
  vision}, pages 1440--1448, 2015.

\bibitem{gan}
Ian Goodfellow, Jean Pouget-Abadie, Mehdi Mirza, Bing Xu, David Warde-Farley,
  Sherjil Ozair, Aaron Courville, and Yoshua Bengio.
\newblock Generative adversarial nets.
\newblock {\em Advances in neural information processing systems}, 27, 2014.

\bibitem{haliassos2021lips}
Alexandros Haliassos, Konstantinos Vougioukas, Stavros Petridis, and Maja
  Pantic.
\newblock Lips don't lie: A generalisable and robust approach to face forgery
  detection.
\newblock In {\em Proceedings of the IEEE/CVF conference on computer vision and
  pattern recognition}, pages 5039--5049, 2021.

\bibitem{he2020momentum}
Kaiming He, Haoqi Fan, Yuxin Wu, Saining Xie, and Ross Girshick.
\newblock Momentum contrast for unsupervised visual representation learning.
\newblock In {\em Proceedings of the IEEE/CVF conference on computer vision and
  pattern recognition}, pages 9729--9738, 2020.

\bibitem{resnet}
Kaiming He, Xiangyu Zhang, Shaoqing Ren, and Jian Sun.
\newblock Deep residual learning for image recognition.
\newblock In {\em Proceedings of the IEEE conference on computer vision and
  pattern recognition}, pages 770--778, 2016.

\bibitem{hernandez2020deepfakeson}
Javier Hernandez-Ortega, Ruben Tolosana, Julian Fierrez, and Aythami Morales.
\newblock Deepfakeson-phys: Deepfakes detection based on heart rate estimation.
\newblock {\em arXiv preprint arXiv:2010.00400}, 2020.

\bibitem{hsu2018learning}
Chih-Chung Hsu, Chia-Yen Lee, and Yi-Xiu Zhuang.
\newblock Learning to detect fake face images in the wild.
\newblock In {\em 2018 International Symposium on Computer, Consumer and
  Control (IS3C)}, pages 388--391. IEEE, 2018.

\bibitem{densenet}
Gao Huang, Zhuang Liu, Laurens van~der Maaten, and Kilian~Q. Weinberger.
\newblock Densely connected convolutional networks.
\newblock In {\em 2017 {IEEE} Conference on Computer Vision and Pattern
  Recognition, {CVPR} 2017, Honolulu, HI, USA, July 21-26, 2017}, pages
  2261--2269. {IEEE} Computer Society, 2017.

\bibitem{LFW}
Gary~B Huang, Marwan Mattar, Tamara Berg, and Eric Learned-Miller.
\newblock Labeled faces in the wild: A database forstudying face recognition in
  unconstrained environments.
\newblock In {\em Workshop on faces in'Real-Life'Images: detection, alignment,
  and recognition}, 2008.

\bibitem{stylegan}
Tero Karras, Samuli Laine, and Timo Aila.
\newblock A style-based generator architecture for generative adversarial
  networks.
\newblock In {\em Proceedings of the IEEE/CVF Conference on Computer Vision and
  Pattern Recognition}, pages 4401--4410, 2019.

\bibitem{analystylegan}
Tero Karras, Samuli Laine, Miika Aittala, Janne Hellsten, Jaakko Lehtinen, and
  Timo Aila.
\newblock Analyzing and improving the image quality of stylegan.
\newblock In {\em Proceedings of the IEEE/CVF Conference on Computer Vision and
  Pattern Recognition}, pages 8110--8119, 2020.

\bibitem{adam}
Diederik~P Kingma and Jimmy Ba.
\newblock Adam: A method for stochastic optimization.
\newblock {\em arXiv preprint arXiv:1412.6980}, 2014.

\bibitem{faceswap}
Marek Kowalski.
\newblock {FaceSwap}.
\newblock \url{https://github.com/MarekKowalski/FaceSwap}, 2018.

\bibitem{Authors14}
FirstName LastName.
\newblock The frobnicatable foo filter, 2014.
\newblock Face and Gesture submission ID 324. Supplied as supplemental material
  {\tt fg324.pdf}.

\bibitem{Authors14b}
FirstName LastName.
\newblock Frobnication tutorial, 2014.
\newblock Supplied as supplemental material {\tt tr.pdf}.

\bibitem{single-center-loss}
Jiaming Li, Hongtao Xie, Jiahong Li, Zhongyuan Wang, and Yongdong Zhang.
\newblock Frequency-aware discriminative feature learning supervised by
  single-center loss for face forgery detection.
\newblock In {\em Proceedings of the IEEE/CVF Conference on Computer Vision and
  Pattern Recognition}, pages 6458--6467, 2021.

\bibitem{faceshifter}
Lingzhi Li, Jianmin Bao, Hao Yang, Dong Chen, and Fang Wen.
\newblock Faceshifter: Towards high fidelity and occlusion aware face swapping.
\newblock {\em arXiv preprint arXiv:1912.13457}, 2019.

\bibitem{face-x-ray}
Lingzhi Li, Jianmin Bao, Ting Zhang, Hao Yang, Dong Chen, Fang Wen, and Baining
  Guo.
\newblock Face x-ray for more general face forgery detection.
\newblock In {\em Proceedings of the IEEE/CVF Conference on Computer Vision and
  Pattern Recognition}, pages 5001--5010, 2020.

\bibitem{li2018ictu}
Yuezun Li, Ming-Ching Chang, and Siwei Lyu.
\newblock In ictu oculi: Exposing ai created fake videos by detecting eye
  blinking.
\newblock In {\em 2018 IEEE International Workshop on Information Forensics and
  Security (WIFS)}, pages 1--7. IEEE, 2018.

\bibitem{li2019scale}
Yanghao Li, Yuntao Chen, Naiyan Wang, and Zhaoxiang Zhang.
\newblock Scale-aware trident networks for object detection.
\newblock In {\em Proceedings of the IEEE/CVF International Conference on
  Computer Vision}, pages 6054--6063, 2019.

\bibitem{li2018exposing}
Yuezun Li and Siwei Lyu.
\newblock Exposing deepfake videos by detecting face warping artifacts.
\newblock {\em arXiv preprint arXiv:1811.00656}, 2018.

\bibitem{celeb}
Yuezun Li, Xin Yang, Pu Sun, Honggang Qi, and Siwei Lyu.
\newblock Celeb-df: A large-scale challenging dataset for deepfake forensics.
\newblock In {\em Proceedings of the IEEE/CVF Conference on Computer Vision and
  Pattern Recognition}, pages 3207--3216, 2020.

\bibitem{spatial}
Honggu Liu, Xiaodan Li, Wenbo Zhou, Yuefeng Chen, Yuan He, Hui Xue, Weiming
  Zhang, and Nenghai Yu.
\newblock Spatial-phase shallow learning: rethinking face forgery detection in
  frequency domain.
\newblock In {\em Proceedings of the IEEE/CVF Conference on Computer Vision and
  Pattern Recognition}, pages 772--781, 2021.

\bibitem{stgan}
Ming Liu, Yukang Ding, Min Xia, Xiao Liu, Errui Ding, Wangmeng Zuo, and Shilei
  Wen.
\newblock Stgan: A unified selective transfer network for arbitrary image
  attribute editing.
\newblock In {\em Proceedings of the IEEE/CVF Conference on Computer Vision and
  Pattern Recognition}, pages 3673--3682, 2019.

\bibitem{liu2018receptive}
Songtao Liu, Di Huang, et~al.
\newblock Receptive field block net for accurate and fast object detection.
\newblock In {\em Proceedings of the European conference on computer vision
  (ECCV)}, pages 385--400, 2018.

\bibitem{ssd}
Wei Liu, Dragomir Anguelov, Dumitru Erhan, Christian Szegedy, Scott Reed,
  Cheng-Yang Fu, and Alexander~C Berg.
\newblock Ssd: Single shot multibox detector.
\newblock In {\em European conference on computer vision}, pages 21--37.
  Springer, 2016.

\bibitem{luo2016understanding}
Wenjie Luo, Yujia Li, Raquel Urtasun, and Richard Zemel.
\newblock Understanding the effective receptive field in deep convolutional
  neural networks.
\newblock {\em Advances in neural information processing systems}, 29, 2016.

\bibitem{marra2018detection}
Francesco Marra, Diego Gragnaniello, Davide Cozzolino, and Luisa Verdoliva.
\newblock Detection of gan-generated fake images over social networks.
\newblock In {\em 2018 IEEE Conference on Multimedia Information Processing and
  Retrieval (MIPR)}, pages 384--389. IEEE, 2018.

\bibitem{masi2020two}
Iacopo Masi, Aditya Killekar, Royston~Marian Mascarenhas, Shenoy~Pratik
  Gurudatt, and Wael AbdAlmageed.
\newblock Two-branch recurrent network for isolating deepfakes in videos.
\newblock In {\em European Conference on Computer Vision}, pages 667--684.
  Springer, 2020.

\bibitem{ExploitingVA}
Falko Matern, C. Riess, and M. Stamminger.
\newblock Exploiting visual artifacts to expose deepfakes and face
  manipulations.
\newblock {\em 2019 IEEE Winter Applications of Computer Vision Workshops
  (WACVW)}, pages 83--92, 2019.

\bibitem{mo2018fake}
Huaxiao Mo, Bolin Chen, and Weiqi Luo.
\newblock Fake faces identification via convolutional neural network.
\newblock In {\em Proceedings of the 6th ACM Workshop on Information Hiding and
  Multimedia Security}, pages 43--47, 2018.

\bibitem{nguyen2019multi}
Huy~H Nguyen, Fuming Fang, Junichi Yamagishi, and Isao Echizen.
\newblock Multi-task learning for detecting and segmenting manipulated facial
  images and videos.
\newblock {\em arXiv preprint arXiv:1906.06876}, 2019.

\bibitem{nguyen2019use}
Huy~H Nguyen, Junichi Yamagishi, and Isao Echizen.
\newblock Use of a capsule network to detect fake images and videos.
\newblock {\em arXiv preprint arXiv:1910.12467}, 2019.

\bibitem{nirkin2021deepfake}
Yuval Nirkin, Lior Wolf, Yosi Keller, and Tal Hassner.
\newblock Deepfake detection based on discrepancies between faces and their
  context.
\newblock {\em IEEE Transactions on Pattern Analysis and Machine Intelligence},
  2021.

\bibitem{poisson}
Patrick P{\'e}rez, Michel Gangnet, and Andrew Blake.
\newblock Poisson image editing.
\newblock In {\em ACM SIGGRAPH 2003 Papers}, pages 313--318. 2003.

\bibitem{ThinkingIF}
Yuyang Qian, Guojun Yin, Lu Sheng, Zixuan Chen, and Jing Shao.
\newblock Thinking in frequency: Face forgery detection by mining
  frequency-aware clues.
\newblock In {\em ECCV}, 2020.

\bibitem{qian2020thinking}
Yuyang Qian, Guojun Yin, Lu Sheng, Zixuan Chen, and Jing Shao.
\newblock Thinking in frequency: Face forgery detection by mining
  frequency-aware clues.
\newblock In {\em European Conference on Computer Vision}, pages 86--103.
  Springer, 2020.

\bibitem{quan2018distinguishing}
Weize Quan, Kai Wang, Dong-Ming Yan, and Xiaopeng Zhang.
\newblock Distinguishing between natural and computer-generated images using
  convolutional neural networks.
\newblock {\em IEEE Transactions on Information Forensics and Security},
  13(11):2772--2787, 2018.

\bibitem{rahmouni2017distinguishing}
Nicolas Rahmouni, Vincent Nozick, Junichi Yamagishi, and Isao Echizen.
\newblock Distinguishing computer graphics from natural images using
  convolution neural networks.
\newblock In {\em 2017 IEEE Workshop on Information Forensics and Security
  (WIFS)}, pages 1--6. IEEE, 2017.

\bibitem{yolov1_confidence_loc_loss}
Joseph Redmon, Santosh Divvala, Ross Girshick, and Ali Farhadi.
\newblock You only look once: Unified, real-time object detection.
\newblock In {\em Proceedings of the IEEE conference on computer vision and
  pattern recognition}, pages 779--788, 2016.

\bibitem{yolov2_anchor}
Joseph Redmon and Ali Farhadi.
\newblock Yolo9000: better, faster, stronger.
\newblock In {\em Proceedings of the IEEE conference on computer vision and
  pattern recognition}, pages 7263--7271, 2017.

\bibitem{yolov3_anchor}
Joseph Redmon and Ali Farhadi.
\newblock Yolov3: An incremental improvement.
\newblock {\em arXiv preprint arXiv:1804.02767}, 2018.

\bibitem{ronneberger2015u}
Olaf Ronneberger, Philipp Fischer, and Thomas Brox.
\newblock U-net: Convolutional networks for biomedical image segmentation.
\newblock In {\em International Conference on Medical image computing and
  computer-assisted intervention}, pages 234--241. Springer, 2015.

\bibitem{ff++}
Andreas Rossler, Davide Cozzolino, Luisa Verdoliva, Christian Riess, Justus
  Thies, and Matthias Nie{\ss}ner.
\newblock Faceforensics++: Learning to detect manipulated facial images.
\newblock In {\em Proceedings of the IEEE/CVF International Conference on
  Computer Vision}, pages 1--11, 2019.

\bibitem{selvaraju2017grad}
Ramprasaath~R Selvaraju, Michael Cogswell, Abhishek Das, Ramakrishna Vedantam,
  Devi Parikh, and Dhruv Batra.
\newblock Grad-cam: Visual explanations from deep networks via gradient-based
  localization.
\newblock In {\em Proceedings of the IEEE international conference on computer
  vision}, pages 618--626, 2017.

\bibitem{shapley-value}
Lloyd~S Shapley.
\newblock A value for n-person games, contributions to the theory of games, 2,
  307--317, 1953.

\bibitem{SBI}
Kaede Shiohara and Toshihiko Yamasaki.
\newblock Detecting deepfakes with self-blended images.
\newblock In {\em Proceedings of the IEEE/CVF Conference on Computer Vision and
  Pattern Recognition}, pages 18720--18729, 2022.

\bibitem{song2022adaptive}
Luchuan Song, Zheng Fang, Xiaodan Li, Xiaoyi Dong, Zhenchao Jin, Yuefeng Chen,
  and Siwei Lyu.
\newblock Adaptive face forgery detection in cross domain.
\newblock In {\em Computer Vision--ECCV 2022: 17th European Conference, Tel
  Aviv, Israel, October 23--27, 2022, Proceedings, Part XXXIV}, pages 467--484.
  Springer, 2022.

\bibitem{song2022face}
Luchuan Song, Xiaodan Li, Zheng Fang, Zhenchao Jin, YueFeng Chen, and Chenliang
  Xu.
\newblock Face forgery detection via symmetric transformer.
\newblock In {\em Proceedings of the 30th ACM International Conference on
  Multimedia}, pages 4102--4111, 2022.

\bibitem{hrnet}
Ke Sun, Bin Xiao, Dong Liu, and Jingdong Wang.
\newblock Deep high-resolution representation learning for human pose
  estimation.
\newblock In {\em Proceedings of the IEEE/CVF Conference on Computer Vision and
  Pattern Recognition}, pages 5693--5703, 2019.

\bibitem{improving}
Zekun Sun, Yujie Han, Zeyu Hua, Na Ruan, and Weijia Jia.
\newblock Improving the efficiency and robustness of deepfakes detection
  through precise geometric features.
\newblock In {\em Proceedings of the IEEE/CVF Conference on Computer Vision and
  Pattern Recognition}, pages 3609--3618, 2021.

\bibitem{efficientnet}
Mingxing Tan and Quoc Le.
\newblock Efficientnet: Rethinking model scaling for convolutional neural
  networks.
\newblock In {\em International Conference on Machine Learning}, pages
  6105--6114. PMLR, 2019.

\bibitem{tariq2018detecting}
Shahroz Tariq, Sangyup Lee, Hoyoung Kim, Youjin Shin, and Simon~S Woo.
\newblock Detecting both machine and human created fake face images in the
  wild.
\newblock In {\em Proceedings of the 2nd international workshop on multimedia
  privacy and security}, pages 81--87, 2018.

\bibitem{neural-textural}
Justus Thies, Michael Zollh{\"o}fer, and Matthias Nie{\ss}ner.
\newblock Deferred neural rendering: Image synthesis using neural textures.
\newblock {\em ACM Transactions on Graphics (TOG)}, 38(4):1--12, 2019.

\bibitem{face2face}
Justus Thies, Michael Zollhofer, Marc Stamminger, Christian Theobalt, and
  Matthias Nie{\ss}ner.
\newblock Face2face: Real-time face capture and reenactment of rgb videos.
\newblock In {\em Proceedings of the IEEE conference on computer vision and
  pattern recognition}, pages 2387--2395, 2016.

\bibitem{van2008visualizing}
Laurens Van~der Maaten and Geoffrey Hinton.
\newblock Visualizing data using t-sne.
\newblock {\em Journal of machine learning research}, 9(11), 2008.

\bibitem{transformer}
Ashish Vaswani, Noam Shazeer, Niki Parmar, Jakob Uszkoreit, Llion Jones,
  Aidan~N. Gomez, Lukasz Kaiser, and Illia Polosukhin.
\newblock Attention is all you need.
\newblock In Isabelle Guyon, Ulrike von Luxburg, Samy Bengio, Hanna~M. Wallach,
  Rob Fergus, S.~V.~N. Vishwanathan, and Roman Garnett, editors, {\em Advances
  in Neural Information Processing Systems 30: Annual Conference on Neural
  Information Processing Systems 2017, December 4-9, 2017, Long Beach, CA,
  {USA}}, pages 5998--6008, 2017.

\bibitem{wang2021m2tr}
Junke Wang, Zuxuan Wu, Jingjing Chen, and Yu-Gang Jiang.
\newblock M2tr: Multi-modal multi-scale transformers for deepfake detection.
\newblock {\em arXiv preprint arXiv:2104.09770}, 2021.

\bibitem{dssim}
Zhou Wang, Alan~C Bovik, Hamid~R Sheikh, and Eero~P Simoncelli.
\newblock Image quality assessment: from error visibility to structural
  similarity.
\newblock {\em IEEE transactions on image processing}, 13(4):600--612, 2004.

\bibitem{xuan2019generalization}
Xinsheng Xuan, Bo Peng, Wei Wang, and Jing Dong.
\newblock On the generalization of gan image forensics.
\newblock In {\em Chinese conference on biometric recognition}, pages 134--141.
  Springer, 2019.

\bibitem{yang2019exposing}
Xin Yang, Yuezun Li, and Siwei Lyu.
\newblock Exposing deep fakes using inconsistent head poses.
\newblock In {\em ICASSP 2019-2019 IEEE International Conference on Acoustics,
  Speech and Signal Processing (ICASSP)}, pages 8261--8265. IEEE, 2019.

\bibitem{zhang2021interpreting}
Hao Zhang, Yichen Xie, Longjie Zheng, Die Zhang, and Quanshi Zhang.
\newblock Interpreting multivariate shapley interactions in dnns.
\newblock In {\em Proceedings of the AAAI Conference on Artificial
  Intelligence}, volume~35, pages 10877--10886, 2021.

\bibitem{mtcnn}
Kaipeng Zhang, Zhanpeng Zhang, Zhifeng Li, and Yu Qiao.
\newblock Joint face detection and alignment using multitask cascaded
  convolutional networks.
\newblock {\em IEEE Signal Processing Letters}, 23(10):1499--1503, 2016.

\bibitem{zhang2017s3fd}
Shifeng Zhang, Xiangyu Zhu, Zhen Lei, Hailin Shi, Xiaobo Wang, and Stan~Z Li.
\newblock S3fd: Single shot scale-invariant face detector.
\newblock In {\em Proceedings of the IEEE international conference on computer
  vision}, pages 192--201, 2017.

\bibitem{multi}
Hanqing Zhao, Wenbo Zhou, Dongdong Chen, Tianyi Wei, Weiming Zhang, and Nenghai
  Yu.
\newblock Multi-attentional deepfake detection.
\newblock In {\em Proceedings of the IEEE/CVF Conference on Computer Vision and
  Pattern Recognition}, pages 2185--2194, 2021.

\bibitem{amazon}
Tianchen Zhao, Xiang Xu, Mingze Xu, Hui Ding, Yuanjun Xiong, and Wei Xia.
\newblock Learning self-consistency for deepfake detection.
\newblock In {\em Proceedings of the IEEE/CVF international conference on
  computer vision}, pages 15023--15033, 2021.

\bibitem{zhou2016learning}
Bolei Zhou, Aditya Khosla, Agata Lapedriza, Aude Oliva, and Antonio Torralba.
\newblock Learning deep features for discriminative localization.
\newblock In {\em Proceedings of the IEEE conference on computer vision and
  pattern recognition}, pages 2921--2929, 2016.

\end{thebibliography}
}

\end{document}